\pdfoutput=1
\documentclass[11pt]{article}

\usepackage{EMNLP2023}
\usepackage{amsmath, amssymb}
\usepackage{times}
\usepackage{latexsym}
\usepackage{graphicx}
\usepackage[T1]{fontenc}
\usepackage[utf8]{inputenc}
\usepackage{microtype}
\usepackage{inconsolata}
\usepackage{hyperref}
\usepackage{multirow}
\usepackage{float}

\title{OpinSummEval: Revisiting Automated Evaluation for Opinion Summarization}

\author{Yuchen Shen \\
  Language Technologies Institute \\
  Carnegie Mellon University \\
  \href{mailto:yuchens3@andrew.cmu.edu}{\textcolor{black}{\texttt{yuchens3@andrew.cmu.edu}}} \\\And
  Xiaojun Wan \\
  Wangxuan Institute of Computer Technology\\
  Peking University \\
  \href{mailto:wanxiaojun@pku.edu.cn}{\textcolor{black}{\texttt{wanxiaojun@pku.edu.cn}}} \\}

\begin{document}
\maketitle
\begin{abstract}
Opinion summarization sets itself apart from other types of summarization tasks due to its distinctive focus on aspects and sentiments. Although certain automated evaluation methods like ROUGE have gained popularity, we have found them to be unreliable measures for assessing the quality of opinion summaries. In this paper, we present \textsc{OpinSummEval}, a dataset comprising human judgments and outputs from 14 opinion summarization models. We further explore the correlation between 26 automatic metrics and human ratings across four dimensions. Our findings indicate that metrics based on neural networks generally outperform non-neural ones. However, even metrics built on powerful backbones, such as BART and GPT-3/3.5, do not consistently correlate well across all dimensions, highlighting the need for advancements in automated evaluation methods for opinion summarization. The code and data are publicly available at \href{https://github.com/A-Chicharito-S/OpinSummEval/tree/main}{https://github.com/A-Chicharito-S/OpinSummEval/tree/main}.
\end{abstract}
\section{Introduction}
Opinion summarization has garnered significant research interest in light of recent advancements in neural networks and large datasets \cite{brazinskas-etal-2020-unsupervised, Amplayo_Angelidis_Lapata_2021}. In contrast to conventional summarization tasks, which focus on preserving key information in unstructured texts like news articles, opinion summarization places emphasis on extracting prevalent aspects and expressing coherent sentiments from a vast number of reviews, which are often disorganized and occasionally contradictory (Table~\ref{tab:unstruct_confict_exm}). Due to the large size of datasets and extensive annotations \cite{pmlr-v97-chu19b}, opinion summarizers \cite{amplayo-lapata-2020-unsupervised, isonuma-etal-2021-unsupervised} are usually trained in an unsupervised manner, where pseudo pairs of \{reviews, summary\} are constructed from a collection of reviews without relying on human-written references \cite{brazinskas-etal-2020-unsupervised}.
\begin{table}[t]
    \centering
    \small
    \begin{tabular}{p{7.2cm}}
    \hline
         \textbf{Review 1}: \\
         Great wash, \textcolor[RGB]{252,76,27}{friendly staff}, \textcolor{red}{can't beat the price}. Get a quality wash for \$5 in 5 minutes. The higher priced washes are good too. Not much more to say than that.  \\
         \hline
         \textbf{Review 2}: \\
         What happened to this place? The owner is a good man, treats people fair, \textit{as a matter fact I once drop my wallet called an hour later, he found it and waited 30 minutes past closing for me to get back and pick it up!} \textcolor[RGB]{252,76,27}{However, recently he \textit{seem} to have hired a bunch of retards who can't understand the concept of service.} But then again, it's \$10 bucks \& location is prime. I'll go back there, but only in a pinch! \\
         \hline
          \textbf{Review 8}: \\
          Pro: Brightly lit, open late;
          Con: \textcolor{red}{Waaay overpriced} unless you typically drive in the mud and need lots of car washes for a monthly fee.\\
    \hline
    \end{tabular}
    \caption{Some reviews in test case 1 from Yelp \cite{pmlr-v97-chu19b}, where \textit{italics} denotes disorganized phrases and \textcolor{red}{col}\textcolor[RGB]{252,76,27}{ors} mark conflicting opinions.}
    \label{tab:unstruct_confict_exm}
\end{table}
\begin{table*}[t]
\centering 
\small
\begin{tabular}{p{15.6cm}}
\hline
\textbf{Reference}\\
Like most places, when this first opened it was excellent. Slowly though it has been going downhill and quality has been getting worse. \textcolor{red}{The gyro meat doesn't seem fresh} and kind of weird looking. I'd say avoid the gyros because they are pretty bad, but their other food is pretty good. \textcolor[RGB]{252,76,27}{Service is ok} too. I was a little disappointed because I expected better from a family run business. \\
\hline
\textbf{Summary}\\
I love pita cafe. The food is delicious and \textcolor[RGB]{252,76,27}{the service is great}. I've had the gyro \textit{and the gyro meat and} \textcolor{red}{\textit{the gyro meat} is fresh}. the gyro meat is a bit dry, but it's still good. I'll definitely be back.\\
\hline
\textbf{R1} / \textbf{avr-R1}: 34.48 / 27.11 \qquad\qquad \textbf{R2} / \textbf{avr-R2}: 8.77 / 4.23 \qquad\qquad \textbf{RL} / \textbf{avr-RL}: 18.96 / 17.83\\
\hline
\end{tabular}
\caption{\label{tab:exm_case_output}
One exemplary summary generated by T5 \cite{JMLR:v21:20-074} with ROUGE-1/2/L exceeding the model average over Yelp. We denote disorganized phrases with \textit{italics} and mark inconsistency with \textcolor{red}{col}\textcolor[RGB]{252,76,27}{ors}.}
\end{table*}
Despite significant advancements in datasets and architectures, evaluating the performance of models for opinion summarization remains a challenge. One common approach is automated evaluation, which employs automatic metrics like ROUGE \cite{lin-2004-rouge} as criteria. While this method is efficient and provides stable results, it may not necessarily accurately reflect the model's performance from a human perspective (Table~\ref{tab:exm_case_output}). Another approach is human evaluation, where annotators are tasked with scoring or ranking summaries from different models. Human evaluation is more closely aligned with common understandings and is therefore considered more reliable than automated scores. However, it is typically time-consuming and labor-intensive, making it suitable primarily for the testing stage and impractical for providing supervision signals during model training.   

Our literature review of 21 papers published between 2018 and 2023 (Appendix~\ref{app_0}), reveals that only 3 papers introduce different metrics as complements to ROUGE for evaluating opinion summarization. We argue that in addition to the advancements made in opinion summarization datasets and models, there should be attention given to the evaluation of metrics in terms of their alignment with human judgments. Such emphasis would be valuable in selecting an appropriate metric that facilitates efficient and human-aligned evaluation of model performance. Moreover, opinion summarization possesses distinctive characteristics, such as its emphasis on aspects, the diversity of opinions and expressions, and the difficulty of expressing coherent sentiments from potentially conflicting reviews. These factors set opinion summarization apart from most other summarization tasks and introduce new challenges for automatic metrics to correlate well with human judgments. Hence, even though certain metrics have demonstrated effectiveness in other summarization tasks \cite{10.1162/tacl_a_00373, gao-wan-2022-dialsummeval}, their reliability and performance in opinion summarization still lack sufficient verification and comprehensive analysis.

Our work is motivated to fill the blank with the following contributions: 1) We introduce \textsc{OpinSummEval}, a dataset with human annotations on the outputs of 14 opinion summarization models over 4 dimensions, which is the first of its kind to the best of our knowledge; 2) We conduct a comprehensive evaluation of 26 metrics for opinion summarization. Our findings indicate that neural-based metrics, such as BARTScore \cite{NEURIPS2021_e4d2b6e6} and ChatGPT \cite{gao2023humanlike}, exhibit superior performance compared to non-neural metrics like ROUGE; 3) We assess the performance of various models (statistically-based, task-agnostic, task-specific, and zero-shot) with \textsc{OpinSummEval}. Our analysis reveals that task-specific models can compensate for the limitations posed by model sizes through specialized paradigms. Furthermore, we observe that GPT-3.5 \cite{bhaskar2023prompted} consistently outperforms other models, as preferred by human evaluators. These contributions collectively enhance our understanding of opinion summarization, provide a benchmark dataset for future research, highlight the effectiveness of neural-based metrics, and offer insights into the performance of different opinion summarization models.
\section{Related Work}
\textbf{Automated Evaluation} Besides the success of metrics \cite{10.3115/1073083.1073135, lin-2004-rouge} that compute n-gram overlaps, such statistically-based measurements usually fail to promote paraphrases that convey the same meaning. Recent advances in automatic metrics \cite{zhao-etal-2019-moverscore, Colombo_Clavel_Piantanida_2022} take insights from neural networks and encourage diversity in words and phrases. \citet{bert-score} propose BERTScore, which measures word-wise similarities with BERT \cite{devlin-etal-2019-bert} embeddings. BARTScore \cite{NEURIPS2021_e4d2b6e6} treats evaluation as a text generation task and uses the conditional probability of BART \cite{lewis-etal-2020-bart} as a metric. \citet{scialom-etal-2019-answers} cast evaluation as a Question Answering (QA) task and measure the quality of texts with a trained QA model. 

As the GPT family raises to power, metrics based on it \cite{wang2023chatgpt, luo2023chatgpt, fu2023gptscore} also show great potential. \citet{gao2023humanlike} instruct ChatGPT to evaluate with an integer score. \citet{liu2023geval} propose to augment the instructions with \textit{Chain of Thought} (CoT) \cite{wei2023chainofthought} and weight a set of predefined integer scores with their generation probabilities from GPT-3/4.

\textbf{Metrics Evaluation in Summarization} \citet{bhandari-etal-2020-evaluating} investigate the effectiveness of metrics for text summarization using \textit{pyramid} \cite{nenkova-passonneau-2004-evaluating}. \citet{10.1162/tacl_a_00373} evaluate metrics in text summarization by annotating the CNN/DailyMail dataset \cite{nallapati-etal-2016-abstractive} in terms of relevance, consistency, fluency, and coherence. \citet{gao-wan-2022-dialsummeval} similarly conduct evaluation for dialogue summarization, and \citet{yuan2023revisiting} evaluate metrics for biomedical question summarization. Similar to our work, \citet{malon2023automatically} constructs ReviewNLI and evaluates 4 metrics on opinion prevalence.

However, none of the tasks share the characteristics of opinion summarization, nor do these works evaluate GPT-based metrics, which motivates our work to evaluate automated methods in the task of opinion summarization. 
\section{Preliminaries}
In this section, we introduce the task definition of metric evaluation, the selected summarization models, and the automatic metrics to be evaluated.
\subsection{Task Definition}
Given a dataset $D$ containing $N$ instances, we denote the $i$-th instance as $d_i$. With $M$ summarization models, we denote $\hat{s}_{j}^i$ as the output from the $j$-th model on $d_i$, and $\mathcal{M}_k(\hat{s}_{j}^i)$ as the score assigned by metric $\mathcal{M}_k$. If we choose $C$ as the correlation criteria, the relation $\mathcal{R}$ between metric $\mathcal{M}_p$ and $\mathcal{M}_q$ is measured at different levels \cite{bhandari-etal-2020-evaluating}: 

\textbf{System-level correlation}
\begin{align}
    &\mathcal{R}_{sys}(p, q)=C( \notag\\
    &[\frac{1}{N}\sum_{i}\mathcal{M}_p(\hat{s}_{1}^i), ..., \frac{1}{N}\sum_{i}\mathcal{M}_p(\hat{s}_{M}^i)],\notag\\
    &[\frac{1}{N}\sum_{i}\mathcal{M}_q(\hat{s}_{1}^i), ..., \frac{1}{N}\sum_{i}\mathcal{M}_q(\hat{s}_{M}^i)])
\end{align}
where the associated p-value reflects the significance of the correlation $\mathcal{R}_{sys}(p, q)$.

\textbf{Summary-level correlation}
\begin{align}
    &\mathcal{R}_{sum}(p, q)=\frac{1}{N}\sum_i C(\notag\\&[\mathcal{M}_p(\hat{s}_{1}^i), ..., \mathcal{M}_p(\hat{s}_{M}^i)],\notag\\
    &[\mathcal{M}_q(\hat{s}_{1}^i), ..., \mathcal{M}_q(\hat{s}_{M}^i)])
\end{align}
where there is \textbf{no} p-value since the correlations are averaged over the dataset $D$.
\subsection{Summarization Models}
We selected 14 popularly used models\footnote{The detailed introduction and the resources for their outputs are listed in Appendix \ref{app_list_of_selected_models}.} in opinion summarization from 4 categories: statistically-based, task-agnostic, task-specific, and zero-shot. We use the superscript $\textsc{Ext}$ and $\textsc{Abs}$ to denote extractive and abstractive models. 

\textit{\textbf{Statistically-Based}} models rely on linguistic features of reviews and respective statistical results to perform extractive summarization. Models in this category include \textbf{LexRank}$^{\textsc{Ext}}$ \cite{10.5555/1622487.1622501}, \textbf{Opinosis}$^{\textsc{Ext}}$ \cite{ganesan-etal-2010-opinosis}, and \textbf{BertCent}$^{\textsc{Ext}}$ \cite{Amplayo_Angelidis_Lapata_2021}.

\textit{\textbf{Task-Agnostic}} models are pre-trained language models (PLMs) intended to fit multiple tasks. Models in this category are finetuned with suggested hyperparameters to achieve competitive performance. We select \textbf{BART}$^{\textsc{Abs}}$ \cite{lewis-etal-2020-bart}, \textbf{T5}$^{\textsc{Abs}}$ \cite{JMLR:v21:20-074}, and \textbf{PEGASUS}$^{\textsc{Abs}}$ \cite{pmlr-v119-zhang20ae} as our backbones.

\textit{\textbf{Task-Specific}} models are designed specifically for opinion summarization, with objectives and modules that attend to obstacles such as unsupervised training. We choose \textbf{COOP}$^{\textsc{Abs}}$ \cite{iso-etal-2021-convex-aggregation}, \textbf{CopyCat}$^{\textsc{Abs}}$ \cite{brazinskas-etal-2020-unsupervised}, \textbf{DenoiseSum}$^{\textsc{Abs}}$ \cite{amplayo-lapata-2020-unsupervised}, \textbf{MeanSum}$^{\textsc{Abs}}$ \cite{pmlr-v97-chu19b}, \textbf{OpinionDigest}$^{\textsc{Abs}}$ \cite{suhara-etal-2020-opiniondigest}, \textbf{PlanSum}$^{\textsc{Abs}}$ \cite{Amplayo_Angelidis_Lapata_2021}, and \textbf{RecurSum}$^{\textsc{Abs}}$ \cite{isonuma-etal-2021-unsupervised} as the representatives.

\textit{\textbf{Zero-Shot}} models are not trained on any datasets for opinion summarization and are tested directly. We choose \textbf{GPT-3.5}$^{\textsc{Abs}}$ \cite{bhaskar2023prompted}, \texttt{text-davinci-003} in specific,  as the backbone.

\subsection{Evaluation Metrics}
We choose 26 metrics\footnote{The detailed introduction and resources for their implementations are listed in Appendix \ref{app_list_of_eval_metrics}.} to evaluate their effectiveness in opinion summarization. They are categorized into non-GPT and GPT-based, depending on whether they are built upon GPTs.

\textit{\textbf{Non-GPT}} metrics include commonly-used measurements in opinion summarization and popularly evaluated metrics from related works \cite{10.1162/tacl_a_00373, gao-wan-2022-dialsummeval}. We choose the following metrics to evaluate: (\textit{statistically-based}) \textbf{ROUGE} \cite{lin-2004-rouge}, \textbf{BLEU} \cite{10.3115/1073083.1073135}, \textbf{METOR} \cite{banerjee-lavie-2005-meteor}, \textbf{TER} \cite{snover-etal-2006-study}, and \textbf{ChrF} \cite{popovic-2015-chrf}; (\textit{neural-based}) \textbf{BERTScore} \cite{bert-score}, \textbf{BARTScore} \cite{NEURIPS2021_e4d2b6e6}, \textbf{BLANC} \cite{vasilyev-etal-2020-fill}, \textbf{BLEURT} \cite{sellam-etal-2020-bleurt}, \textbf{InfoLM} \cite{Colombo_Clavel_Piantanida_2022}, \textbf{BaryScore} \cite{colombo-etal-2021-automatic}, \textbf{MoverScore} \cite{zhao-etal-2019-moverscore}, \textbf{Sentence Mover’s Similarity} \cite{clark-etal-2019-sentence}, \textbf{EmbeddingAverage} \cite{EmbeddingAvr}, \textbf{VectorExtrema} \cite{forgues2014bootstrapping}, \textbf{GreedyMatching} \cite{rus-lintean-2012-comparison}, \textbf{Perplexity}-[\texttt{PEGASUS}], with PEGASUS as the backbone, \textbf{Prism} \cite{thompson-post-2020-automatic}, \textbf{$\textit{S}^3$} \cite{peyrard-etal-2017-learning} and \textbf{SUPERT} \cite{gao-etal-2020-supert}; (\textit{QA-based}) \textbf{QAFactEval} \cite{fabbri-etal-2022-qafacteval}, \textbf{QuestEval} \cite{scialom-etal-2021-questeval} and \textbf{SummaQA} \cite{scialom-etal-2019-answers}; (\textit{NLI-based}) \textbf{SummaC} \cite{10.1162/tacl_a_00453}.

\textit{\textbf{GPT-Based}} metrics are built upon the GPT family and its variants. Specifically, we choose \textbf{Perplexity}-[\texttt{GPT-2}], with GPT-2 \cite{radford2019language} as the language model, \textbf{ChatGPT} \cite{gao2023humanlike}, with \texttt{gpt-3.5-turbo} as the backbone\footnote{The prompts we use are shown in Appendix \ref{app_prompts_of_chatgpt}.} and two variants\footnote{The prompts and CoT are shown in Appendix \ref{app_prompts_and_CoT_of_GEval}.} of \textbf{G-Eval} \cite{liu2023geval}: \textbf{G-Eval}-[\texttt{text-ada-001}], which weights a set of predefined scores with the generation probabilities conditioned on the instructions and CoT, with \texttt{text-ada-001}\footnote{In \cite{liu2023geval}, the choice is \texttt{text-davinci-003}, a GPT-3.5 variant, which is more powerful however less efficient and more expensive compared with \texttt{text-ada-001}.} as the backbone; \textbf{G-Eval}-[\texttt{gpt-3.5-turbo}], which gives integer scores based on the instructions and CoT, with \texttt{gpt-3.5-turbo} as the scoring model.

\section{\textsc{OpinSummEval}}
In this section, we introduce the dataset upon which annotations are carried out, the 4 dimensions to be annotated, the detailed annotation process, and the analysis of the annotation results.
\subsection{Dataset}
Yelp \cite{pmlr-v97-chu19b} is a widely-used dataset that has promoted vast research works in opinion summarization, upon which the model outputs we are able to collect are the most\footnote{A discussion on such a choice is shown in Appendix \ref{discussion_choice_dataset}}. We base our annotations on its test set, where there are 100 instances and each consists of 8 reviews on the same product/service and 1 human-written reference.

\subsection{Dimensions}
Instead of choosing \textbf{coherence}, \textbf{consistency}, \textbf{fluency}, and \textbf{relevance} \cite{10.1162/tacl_a_00373, gao-wan-2022-dialsummeval} as the dimensions to evaluate, we select the following 4 dimensions consistent with the characteristics of opinion summarization.

\textbf{Aspect Relevance} measures whether the mainly discussed aspects in the reviews are covered exactly by the summary. It focuses on whether the summary correctly reflects the mainly discussed aspects in the reviews.

\textbf{Self-Coherence} measures whether the summary is consistent within itself in terms of sentiments and aspects. It focuses on whether the summary is coherent and does not reflect conflicting opinions.

\textbf{Sentiment Consistency} measures whether the summary is consistent with the reviews in terms of sentiments for each aspect. It focuses on whether the summary aspect-wisely captures the main sentiment in the reviews.

\textbf{Readability} measures whether the summary is fluent and informative. It focuses on whether the summary is well-written and valuable.

\subsection{Process}
The annotation is carried out on the test set of Yelp with the outputs of the aforementioned 14 models. For each instance, we ask the annotators to rate on an integer scale from 1 (worst) to 5 (best) and annotate every summary independently over the 4 dimensions. The overall workload would be 2 (\textit{\# of annotators}) $\times$ 100 (\textit{\# of instances}) $\times$ 14 (\textit{\# of models}) $\times$ 4 (\textit{\# of dimensions}) = 11200 scores, where each dimension receives 2 annotations. 

The annotation is conducted independently and each annotator rates one batch (with a size of 10) at a time to ensure consistency and reliability. The final score of a summary at each dimension is the average of its annotations, and the annotation process with guidelines is detailed in Appendix \ref{app_detailed_annotation_process}.


\subsection{Analysis}
\begin{figure}[t]
    \includegraphics[width=0.5\textwidth]{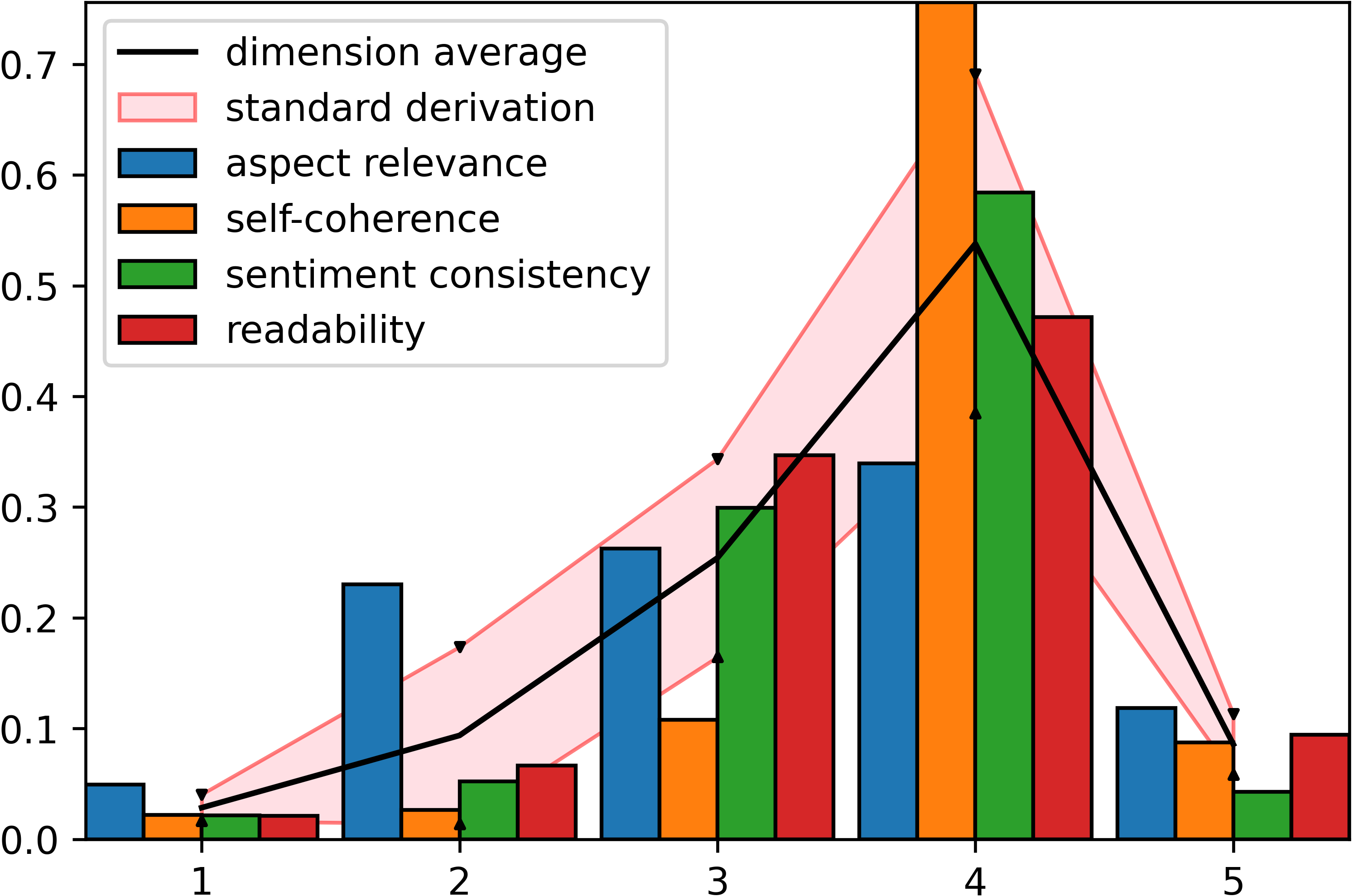}
    \caption{The annotation distribution for each dimension. For each score, we plot the average frequency it is being scored across 4 dimensions, with $\pm$ its standard deviation (marked with $\blacktriangledown$ and $\blacktriangle$).} 
    \label{fig:annot_distri}
\end{figure}
\textbf{Annotation Distribution} We count the annotations for different dimensions\footnote{A sample size of 2 (\textit{\# of annotators}) $\times$ 100 (\textit{\# of instances}) $\times$ 14 (\textit{\# of models}) = 2800 for each.} and show their distributions in Figure~\ref{fig:annot_distri}. 
The dissimilarity of annotation distributions among any two dimensions is evident, suggesting that \textsc{OpinSummEval} maintains independence across dimensions. We observe that the majority of annotations assign a score of 3 or 4 across the four dimensions, which indicates that most models can consistently generate moderately high-quality summaries across various dimensions.
Regarding deviations within each score, we have observed that scores ranging from 2 to 4 exhibit significant variability, whereas scores of 1 and 5 demonstrate relatively smaller deviations. We argue this is due to the fact that summaries evaluated as the worst/best in one dimension often tend to perform poorly/exceptionally across others as well.

\textbf{Annotation Agreement} 
\begin{table}[t]
\centering
\begin{tabular}{lcc}
\hline
\textbf{} & \textbf{Cohen's $\kappa$} & \textbf{Gwet's AC1}\\
\hline
\textbf{Asp.Rel.} & 0.9055 & 0.9217 \\
\textbf{Sel.Coh.} & 0.7788 & 0.9069 \\
\textbf{Sen.Con.} & 0.8295 & 0.9033 \\
\textbf{Readability} & 0.7771 & 0.8537 \\
\hline
\end{tabular}
\caption{\label{tab:agreement_result} The annotation agreement for each dimension.}
\end{table}
We choose Cohen's $\kappa$ \cite{doi:10.1177/001316446002000104} and Gwet's AC1 \cite{https://doi.org/10.1348/000711006X126600} to evaluate the annotation agreement. As shown in Table~\ref{tab:agreement_result}, we report the averaged agreement over the batches for each dimension. We observe that Cohen's $\kappa$ is within an acceptable range\footnote{We show its interpretation and the agreement measured under Fleiss' $\kappa$ \cite{fleiss1971measuring} and Krippendorff’s $\alpha$ \cite{krippendorff2011computing} in Appendix \ref{app_interpretation_of_cohens_kappa}.} between 0.7771 to 0.9055, and the annotators tend to have a higher agreement in terms of ``aspect relevance'' and ``sentiment consistency'' compared with the other two dimensions. This is reasonable since evaluating ``aspect relevance'' and ``sentiment consistency'' involve cross-examination with the reviews, while the others are rated self-referentially. A similar trend is also observed for Gwet's AC1.

\section{Evaluation Results}
\begin{table*}[p]
\centering
\begin{tabular}{|l|cc|cc|cc|cc|}
\hline
    \textbf{} & \multicolumn{2}{c|}{\textbf{Asp.Rel.}} & \multicolumn{2}{c|}{\textbf{Sel.Coh.}} & \multicolumn{2}{c|}{\textbf{Sen.Con.}} & \multicolumn{2}{c|}{\textbf{ Read.}}\\
    \hline
    \textbf{metric} & sys & sum & sys & sum & sys & sum & sys & sum\\
    \hline
    \textbf{ROUGE-1} & -0.12 & 0.11 & -0.23 & 0.09 & -0.29 & 0.00 & -0.02 & 0.06 \\
    \textbf{ROUGE-2} & 0.01 & 0.11 & -0.05 & 0.11 & -0.15 & 0.04 & 0.11 & 0.10 \\
    \textbf{ROUGE-L} & 0.10 & 0.14 & 0.03 & 0.15 & -0.07 & 0.05 & 0.11 & 0.10 \\
    \textbf{BLEU-1} & -0.23 & 0.08 & -0.21 & 0.07 & -0.31 & -0.03 & -0.04 & 0.03 \\
    \textbf{BLEU-2} & -0.12 & 0.11 & -0.19 & 0.11 & -0.29 & 0.04 & -0.07 & 0.07 \\
    \textbf{BLEU-3} & -0.01 & 0.12 & -0.08 & 0.12 & -0.20 & 0.05 & 0.09 & 0.09 \\
    \textbf{BLEU-4} & -0.05 & 0.12 & -0.12 & 0.12 & -0.24 & 0.04 & 0.04 & 0.09 \\
    \textbf{METEOR} & -0.16 & 0.14 & -0.19 & 0.09 & -0.15 & 0.04 & 0.00 & 0.06 \\
    \textbf{TER} & 0.12 & -0.05 & 0.10 & -0.09 & 0.18 & -0.02 & -0.07 & -0.19 \\
    \textbf{ChrF} & -0.01 & 0.15 & 0.01 & 0.12 & 0.00 & 0.05 & 0.07 & 0.05 \\
    \textbf{BERTScore$_{precision}$} & 0.05 & 0.12 & 0.03 & 0.14 & -0.04 & 0.05 & 0.20 & 0.26 \\
    \textbf{BERTScore$_{recall}$} & 0.19 & \textbf{0.20} & 0.25 & \textbf{0.19} & 0.13 & 0.10 & \textbf{0.44}$^*$ & \textbf{0.30} \\
    \textbf{BERTScore$_{f1}$} & 0.19 & 0.16 & 0.16 & 0.17 & 0.00 & 0.06 & 0.38 & \textbf{0.30} \\
    \textbf{BARTScore$_{hyp\rightarrow ref}$} & 0.41$^*$ & \textbf{0.19} & 0.43$^*$ & 0.12 & 0.31 & 0.12 & 0.42$^*$ & 0.10 \\
    \textbf{BARTScore$_{ref\rightarrow hyp}$} & 0.25 & 0.17 & 0.23 & \textbf{0.17} & 0.15 & 0.08 & 0.31 & 0.22 \\
    \textbf{BARTScore$_{rev\rightarrow hyp}$}$^\blacktriangledown$ & \textcolor{red}{\textbf{0.65}}$^{**}$ & \textbf{0.22} & \textcolor{red}{\textbf{0.76}}$^{**}$ & \textcolor{red}{\textbf{0.29}} & \textcolor{red}{\textbf{0.77}}$^{**}$ & \textbf{0.34} & \textbf{0.46}$^*$ & \textbf{0.33} \\
    \textbf{BLANC$_{help}$}$^\blacktriangledown$ & \textbf{0.56}$^{**}$ & 0.17 & 0.54$^{**}$ & 0.16 & \textbf{0.62}$^{**}$ & \textbf{0.24} & 0.38 & 0.17 \\
    \textbf{BLANC$_{tune}$}$^\blacktriangledown$ & 0.49$^*$ & 0.14 & 0.47$^*$ & 0.10 & 0.55$^{**}$ & 0.19 & 0.31 & 0.09 \\
    \textbf{BLEURT} & 0.38 & \textbf{0.21} & 0.36 & \textbf{0.22} & 0.29 & 0.16 & \textbf{0.53}$^{**}$ & \textbf{0.30} \\
    \textbf{InfoLM} & 0.25 & -0.08 & 0.19 & -0.01 & 0.20 & 0.00 & 0.18 & 0.00 \\
    \textbf{BaryScore} & 0.10 & -0.14 & 0.16 & -0.11 & 0.27 & 0.01 & 0.00 & -0.13 \\
    \textbf{MoverScore} & -0.10 & 0.14 & -0.16 & 0.10 & -0.27 & -0.01 & 0.00 & 0.12 \\
    \textbf{SMS$_{ELMo}$} & 0.23 & 0.11 & 0.30 & 0.11 & 0.27 & 0.07 & 0.35 & 0.14 \\
    \textbf{SMS$_{GLoVe}$} & 0.27 & 0.11 & 0.25 & 0.08 & 0.38 & 0.08 & 0.31 & 0.10 \\
    \textbf{PPL}-[\texttt{PEGASUS}]$^\blacktriangledown$ & -0.08 & -0.06 & -0.01 & -0.07 & 0.02 & -0.03 & -0.22 & -0.07 \\
    \textbf{SUPERT}$^\blacktriangledown$ & \textbf{0.54}$^{**}$ & 0.17 & \textbf{0.56}$^{**}$ & 0.14 & \textbf{0.60}$^{**}$ & 0.18 & 0.40$^*$ & 0.12 \\
    \textbf{QAFactEval}$^\blacktriangledown$ & 0.45$^*$ & 0.08 & 0.47$^*$ & 0.09 & 0.51$^*$ & \textbf{0.21} & 0.27 & 0.13 \\
    \textbf{QuestEval} & 0.43$^*$ & 0.16 & 0.45$^*$ & 0.13 & 0.49$^*$ & \textbf{0.22} & 0.33 & 0.15 \\
    \textbf{SummaQA$_{fscore}$}$^\blacktriangledown$ & \textbf{0.56}$^{**}$ & 0.10 & \textbf{0.58}$^{**}$ & 0.09 & \textbf{0.66}$^{**}$ & 0.17 & 0.38 & 0.10 \\
    \textbf{SummaQA$_{conf}$}$^\blacktriangledown$ & \textbf{0.58}$^{**}$ & 0.10 & \textbf{0.60}$^{**}$ & 0.12 & \textbf{0.69}$^{**}$ & 0.14 & \textbf{0.44}$^*$ & 0.15 \\
    \textbf{SummaC$_{snt}$}$^\blacktriangledown$ & 0.19 & -0.00 & 0.16 & 0.01 & 0.22 & 0.12 & -0.04 & 0.01 \\
    \textbf{SummaC$_{doc}$}$^\blacktriangledown$ & 0.30 & 0.06 & 0.27 & 0.01 & 0.40$^*$ & 0.17 & 0.11 & 0.05 \\
    \hline
    \textbf{PPL}-[\texttt{GPT-2}]$^\blacktriangledown$ & -0.10 & -0.14 & -0.08 & -0.15 & 0.00 & -0.05 & -0.24 & -0.16 \\
    \textbf{G-Eval}-[\texttt{text-ada-001}]$^\blacktriangledown$ & -0.01 & -0.01 & -0.05 & -0.02 & 0.22 & 0.08 & 0.27 & 0.08 \\
    \textbf{G-Eval}-[\texttt{text-ada-001}]-n$^\blacktriangledown$ & 0.05 & 0.01 & 0.12 & 0.14 & 0.40$^*$ & 0.07 & 0.29 & 0.06 \\
    \textbf{G-Eval}-[\texttt{gpt-3.5-turbo}]$^\blacktriangledown$ & 0.45$^*$ & \textbf{0.23} & \textbf{0.56}$^{**}$ & \textbf{0.26} & 0.55$^{**}$ & \textcolor{red}{\textbf{0.34}} & \textbf{0.53}$^{**}$ & \textbf{0.36} \\
    \textbf{ChatGPT}-[\texttt{gpt-3.5-turbo}]$^\blacktriangledown$ & \textbf{0.56}$^{**}$ & \textcolor{red}{\textbf{0.30}} & \textbf{0.62}$^{**}$ & \textbf{0.25} & \textbf{0.56}$^{**}$ & \textbf{0.33} & \textcolor{red}{\textbf{0.62}}$^{**}$ & \textcolor{red}{\textbf{0.42}} \\
\hline
\end{tabular}
\caption{\label{tab:metric_eval_result} The Kendall's $\tau$ correlations at system-level and summary-level between automatic metrics and human annotations over 4 dimensions. The best and 2$^{nd}$- to 6$^{th}$-best systems are respectively marked in \textcolor{red}{\textbf{red}} and \textbf{black}. $*$ and $**$ denote a p-value of $\leq 0.05$ and $\leq 0.01$. The superscript $\blacktriangledown$ marks metrics that evaluate w/o references. Inside the brackets [X] denotes the backbone model X used for the metric. For \textbf{BARTScore}, $hyp$, $ref$, and $rev$ stand for model summary, reference summary, and input reviews, where $A\rightarrow B$ computes the generation probability from $A$ to $B$. For \textbf{BLANC}, $help$ measures the difference in accuracy between reconstructions of the summary and a ``filler'', and $tune$ refers to that between the tuned model and the original model. \textbf{SMS} is the abbreviation for Sentence Mover's Similarity and \textbf{PPL} stands for Perplexity. For \textbf{SummaQA}, $fscore$ reflects the average overlaps between the predicted and ground-truth answers, and $conf$ stands for the confidence of the prediction. \textit{snt} and \textit{doc} stand for sentence- / document-level evaluation for \textbf{SummaC}, respectively. The ``n'' in \textbf{G-Eval}-[\texttt{text-ada-001}]-n stands for normalization, where the weights for a set of predefined scores are normalized to sum up to 1.}
\end{table*}
\subsection{Metric Evaluation}\label{metric_eval}
We measure the correlations between metrics and human annotations with Kendall's $\tau$\footnote{A discussion of Pearson's r is detailed in Appendix~\ref{app_discussion_on_pearsons_r}.} following \citet{10.1162/tacl_a_00373} and show the results in Table~\ref{tab:metric_eval_result}. We observe that certain metrics exhibit a stronger correlation at summary-level than at system-level, such as \textbf{ROUGE-1} at sentiment consistency and \textbf{MoverScore} at aspect relevance, which is similar to the findings of \citet{bhandari-etal-2020-evaluating}. However, it is worth noting that for metrics that show a significant correlation (p-value $\leq 0.05$), there is indeed a higher correlation at system-level than at summary-level across all the dimensions.


Our observations reveal that metrics relying on linguistic features, such as n-gram overlaps, exhibit lower correlations with human judgments across all four dimensions when compared to neural automatic metrics. In the case of the \textbf{ROUGE-n} family, it is commonly believed that ROUGE-1/2 assess informativeness, while ROUGE-L measures fluency \cite{amplayo-etal-2021-aspect}. However, despite their popularity in opinion summarization, their performance is rather unsatisfactory. None of them exhibits a high correlation with human evaluations, which is consistent with the findings of \citet{tay-etal-2019-red}. Based on our evaluation results, we recommend exercising caution when using ROUGE scores to provide training supervision or evaluate the quality of opinion summaries during testing. Regarding other statistically-based metrics like \textbf{BLEU}, \textbf{METEOR}, and \textbf{ChrF}, although they exhibit higher absolute correlation values compared to the ROUGE-n family, their correlations tend to be negative at the system-level and positive at the summary-level. This can potentially cause difficulties when interpreting their meanings. The only exception is \textbf{TER}, which demonstrates positive correlations at the system-level across most dimensions. However, the summary-level correlations are reversed, and overall, TER exhibits low and insignificant correlations at both levels.

Metrics based on neural networks generally exhibit strong correlations with human judgments across all four dimensions. Among all the variants, \textbf{BERTScore}$_{recall}$ demonstrates the highest performance. This can be attributed to the fact that the recall score measures the extent to which words in the summary match the reference. This similarity is akin to determining whether important opinions from the reviews (mentioned in the reference) are captured in the summary. We observe \textbf{BARTScore}$_{rev\rightarrow hyp}$ consistently outperforms others across almost all four dimensions. We believe this superiority has two key factors. First, BART's power as a competitive backbone enables the measurement of conditional generation probabilities. Second, \textbf{BARTScore}$_{rev\rightarrow hyp}$ directly measures the likelihood of a summary being generated from input reviews, which aligns with the main concept of summary evaluation. 

Compared to \textbf{SMS}\footnote{\textbf{SMS} is the abbreviation for Sentence Mover's Similarity.}, \textbf{InfoLM}, and \textbf{BaryScore}, whose correlations are relatively low in magnitude and less significant, \textbf{BLANC} treat evaluation as a language understanding task of the input documents, and achieve high correlations with dimensions that involve analyzing the reviews.
Surprisingly, \textbf{BLEURT} exhibits strong correlations with readability and outperforms BLEU, ROUGE, and BERTScore, which are the three signals used in its training. This suggests that trainable metrics that learn from other metrics can yield competitive and even superior results. The competitive performance of \textbf{SUPERT} can be attributed to its pseudo reference, which comprises sentences extracted from reviews. However, since the extracted sentences can vary in style, they may not serve as a reliable proxy for measuring readability. 

For QA-based metrics, \textbf{QAFactEval}, \textbf{QuestEval}, and \textbf{SummaQA} all exhibit good correlations with dimensions reflecting relevance and consistency, in line with the observations of \citet{gao-wan-2022-dialsummeval}. Although \textbf{SummaC} cast evaluation as a natural language inference (NLI) task to measure factual consistency, the correlation is merely salient even in dimensions that reflect consistency. We suspect the reasons are two-fold: 1) self-consistency and sentiment-consistency focus more on the summary itself and sentiments instead of facts between the reviews and the summary, which is shifted from the original purpose of \textbf{SummaC}; 2) opinions that are potentially scattered and heterogeneous in the reviews make it harder for the model to inference correctly; thus, degrade the evaluation results. 

Among GPT-based metrics, \textbf{PPL}-[\texttt{GPT-2}] performs similarly to \textbf{PPL}-[\texttt{PEGASUS}]\footnote{\textbf{PPL} stands for perplexity.} and exhibit poor correlations across all dimensions. For the two variants of \textbf{G-Eval}-[\texttt{text-ada-001}], they exhibit limited alignment with human annotations. We suspect this is because \texttt{text-ada-001} is a faster however less powerful backbone compared to the original choice, \texttt{text-davinci-003}. This suggests that future directions may focus on developing metrics that prioritize efficiency without compromising quality. Regarding GPT-3.5-based metrics, \textbf{ChatGPT}-[\texttt{gpt-3.5-turbo}] with handcrafted prompts generally outperforms CoT-enhanced \textbf{G-Eval}-[\texttt{gpt-3.5-turbo}]. We believe that this gap can be attributed to: 1) differences in the prompts used, 2) an increase in input length after embedding CoT, and 3) potential drawbacks of using CoT without demonstrations, and further investigation is necessary to understand the details. It is worth noting that ChatGPT-based metrics excel in measuring readability, indicating their potential as effective evaluators of linguistic soundness.

We observe that reference-free metrics (marked by $\blacktriangledown$) generally outperform metrics that rely on reference-based evaluation. While the specific reasons require further investigation, we suspect that the evaluation results may be strongly influenced by the quality and style of human-written references. Therefore, future research could also explore the possibility of reference-free evaluation methods.
\begin{table*}[ht]
\centering \small
\resizebox{\textwidth}{!}{\begin{tabular}{l|cccc|ccc|c}
\hline
    \textbf{models} & {\textbf{Asp.Rel.}} & {\textbf{Sel.Coh.}} & {\textbf{Sen.Con.}} & {\textbf{ Read.}} & {\textbf{R1}} & {\textbf{R2}} & {\textbf{RL}} & \textbf{BARTScore}\\
    \hline
    \textbf{LexRank}$_\triangle$ & 3.475 & \textbf{4.165} & \textbf{3.960} & 3.780 & 24.46 & 2.82 & 13.76 & \textcolor{red}{\textbf{-0.440}}\\
    \textbf{Opinosis}$_\triangle$ & 1.970 & 2.980 & 2.720 & 2.520 &  13.41 & 1.32 & 9.55 & -2.874\\
    \textbf{BertCent} & 3.435 & 4.040 & \textbf{3.960} & 3.715 & 26.67 & 3.19 & 14.67 & \textbf{-1.762}\\
    \hline
    \textbf{BART}$_\triangle$ & \textbf{3.675} & 4.095 & 3.765 & \textbf{4.120} & 31.49 & 5.72 & 19.04 & -1.970\\
    \textbf{T5}$_\triangle$ & 3.450 & 4.020 & 3.795 & 3.555 & 27.11 & 4.23 & 17.83 & -1.946\\
    \textbf{PEGASUS}$_\triangle$ & 3.565 & 4.045 & 3.850 & 3.780 & 27.45 & 4.60 & 18.30 & -1.849\\
    \hline
    \textbf{COOP} & 3.405 & 3.945 & 3.560 & 3.865 & \textcolor{red}{\textbf{35.37}} & \textcolor{red}{\textbf{7.35}} & \textcolor{red}{\textbf{19.94}} & -2.051\\
    \textbf{CopyCat} & 3.330 & 3.960 & 3.605 & 3.935 & 29.47 & 5.26 & 18.09 & -2.096\\
    \textbf{DenoiseSum} & 3.145 & 3.535 & 3.450 & 3.070 & 30.14 & 4.99 & 17.65 & -4.019\\
    \textbf{MeanSum} & 2.910 & 3.740 & 3.285 & 3.280 &  28.86 & 3.66 & 15.91 & -3.008\\
    \textbf{OpinionDigest} & 3.135 & 3.860 & 3.405 & 3.025 & 29.30 & 5.77 & 18.56 & -2.586\\
    \textbf{PlanSum} & 3.255 & 3.925 & 3.470 & 3.700 & \textbf{34.79} & \textbf{7.01} & \textbf{19.74} & -2.344\\
    \textbf{RecurSum} & 2.780 & 3.550 & 3.140 & 2.990 & 33.24 & 5.15 & 18.01 & -3.002\\
    \hline
    \textbf{GPT-3.5}$_\triangle$ & \textcolor{red}{\textbf{3.945}} & \textcolor{red}{\textbf{4.185}} & \textcolor{red}{\textbf{4.085}} & \textcolor{red}{\textbf{4.385}} & 26.58 & 4.15 & 16.13 & -1.803\\
\hline
\end{tabular}}
\caption{\label{tab:human_annot_result}Human ratings over 4 dimensions, ROUGE scores, and BARTScores ($rev\rightarrow hyp$) for 14 models on the Yelp dataset. \textcolor{red}{\textbf{Red}} and \textbf{black} respectively mark the best and the second-best system. ROUGE scores for models with subscript $\triangle$ are calculated by ourselves, the others are from their original papers.}
\end{table*}

\subsection{Model Evaluation}\label{model_eval}
We evaluate the performance of the 14 models based on their average scores over the 4 dimensions and present the results in Table~\ref{tab:human_annot_result}. We also report ROUGE-1/2/L scores (by convention) and BARTScore results (based on previous analysis). 

Extractive models (\textbf{LexRank}, \textbf{BertCent}) are favored over all the dimensions since they select salient sentences from the reviews, which are usually informative and grammatically correct, as summaries. The only exception is \textbf{Opinosis}, and we suspect this is because the model extracts incomplete phrases from the reviews and subsequently re-arranges them, which may result in confusing and potentially inaccurate summaries.

In comparison to \textit{task-agnostic} PLMs, the performance of \textit{task-specific} models is not consistently superior across all dimensions. We believe there are two primary reasons for this. First, we use PLMs with a depth of \textit{at least} 12 layers, which is significantly larger than that of the \textit{task-specific} models. Second, the training paradigms used for \textit{task-specific} models may enhance performance in one dimension while potentially hindering it in another. For example, \textbf{OpinonDigest} is trained to reconstruct a sentence based on a set of extracted keywords. While this training approach may promote self-coherence, it can also lead to hallucinations and potential inconsistencies when compared to the reviews. However, it is important to note that our observations do not contradict the effectiveness of their proposed architectures and training schedules. Notably, we observe that \textbf{CopyCat} achieves comparable performance to \textbf{T5} (3.960 vs. 4.020) in terms of self-coherence, and \textbf{COOP} receives a higher rating for readability compared to \textbf{PEGASUS} (3.865 vs. 3.780). This suggests that their paradigms specifically designed for opinion summarization can compensate for the discrepancy in size and yield comparable capabilities. We present case study on model outputs in Appendix~\ref{app_case_study}.

\section{Discussion}
\subsection{The Choice of Metrics}
Despite our work has shown, as demonstrated in many similar research works \cite{10.1162/tacl_a_00373, gao-wan-2022-dialsummeval}, that n-gram-based automated metrics, such as BLEU \cite{10.3115/1073083.1073135} and ROUGE \cite{lin-2004-rouge}, are less aligned with humans compared with the newly-developed neural-based methods\footnote{Here and followed in this subsection, by ``neural-based'' we refer to metrics whose evaluation paradigms involving neural models other than statistical counting such as n-grams.}, such as BARTScore \cite{NEURIPS2021_e4d2b6e6} and G-Eval \cite{liu2023geval}, we would like to suggest that \textit{choosing which metrics to evaluate opinion summarization models remains to be an \textbf{unresolved} issue}. 

On one hand, it is indeed that neural-based methods show higher correlations with human evaluations, however, it is worth mentioning that these methods might be inherently partial, \textit{for example}, there might exist gender or social biases in the embeddings of a pre-trained model that is later used as the backbone of a neural-based metric, which might implicitly favor opinion summarizers that promote such biases. 

On the other hand, despite the fact that n-gram-based metrics provide fast and efficient evaluations for both training and testing, their statistical nature destines that these metrics are hard to capture the rich variations of human languages, and thus, might indirectly favor models that are more aligned to a limited set of human-written summaries, which would be less flexible and thus less likely to satisfy the increasing demand of controllable opinion summarization \cite{amplayo-lapata-2021-informative, amplayo-etal-2021-aspect, hosking-etal-2023-attributable}. 

Apart from the above dilemmas, both statistical-based and neural-based metrics rely on the number and quality of human-written summaries\footnote{This is also true for reference-free metrics that evaluate with some neural models, which are trained in a supervised fashion with human-annotated labels.}, which might largely affect the evaluation outcomes, \textit{for example}, if the maximum length of human-written references is less than $L$, then models producing summaries exceed $L$ are less likely to receive high scores, despite the fact that long texts sometimes could convey more details that might be beneficial for decision making.

Therefore, we suggest that automated metrics should be chosen carefully when used to evaluate the performance of opinion summarization models. Although the results in this work could potentially be a reference to motivate a specific choice, however, we argue that such a decision would be better made if multiple considerations were taken instead of solely based on our analyses, since the reported correlations are not an absolute criterion to show that one metric is universally better than another.
\subsection{Potential Evaluation Paradigms for Opinion Summarization}
Since there are no metrics particularly tailored for opinion summarization at the time of this research, we would like to suggest some potential evaluation paradigms that might be effective for the development of opinion-summarization-specific metrics.

From the analyses in Section \ref{metric_eval}, we can see that QA-based (e.g., SummaQA) and text-generation-alike (e.g., BARTScore) evaluation paradigms could be potential directions to develop novel metrics for opinion summarization, especially with the recent advancement of large language models (LLMs), which show astonishing ability in both QA and language modeling. Comparing the performance of BERTScore and BARTScore, we can also conclude that the training objectives of the backbones affect the final evaluation results; thus, future works could further consider building metrics based on some opinion summarization models, whose training objectives naturally align with the evaluation process.

Based on the evaluation results from Section \ref{model_eval}, we can observe that among \textit{task-specific} models, COOP ranks the best measured by both ROUGE and BARTScore, and is favored by human annotators across different dimensions as well. COOP first searches a convex combination of the latent representations based on input-output word overlaps, and then uses the searched latent vector to produce summaries, which is similar to the best performing automated metric \textbf{BARTScore}$_{rev\rightarrow hyp}$ that evaluates via $(rev, hyp)$ matching. The prominent performance of COOP and its resemblance to \textbf{BARTScore}$_{rev\rightarrow hyp}$ suggest that future works could take inspiration from COOP, and design metrics based on input-output matching to evaluate models for opinion summarization.

\section{Conclusion}


We present \textsc{OpinSummEval}, a dataset that contains summaries from 14 opinion summarization models, annotated across four dimensions. Through a comprehensive investigation and analysis, we have the following findings: 1) Metrics based on n-gram statistics, such as ROUGE, exhibit poor correlations with human evaluation. Therefore, despite their popularity, future works in opinion summarization should be cautious when using these metrics; 2) Neural-based metrics perform better than non-neural metrics. However, it is important to note that the performance of powerful backbone models does not guarantee high correlations with human evaluation; 3) Only a few metrics consistently align well with human evaluation across all four dimensions, and BARTScore and QA-based metrics demonstrate competitive performance across multiple dimensions. This suggests that future development of metrics for opinion summarization could draw inspiration from the paradigms used in these metrics; 4) Recently proposed metrics based on GPT-3/3.5 excel in evaluating readability. However, their performance in other dimensions is influenced by the choice of prompts and backbones. Careful consideration is suggested if these metrics are used for evaluation in opinion summarization.

Based on our research, we hope that future works will recognize the importance of selecting proper evaluation methods, consider using metrics in addition to ROUGE, and even design novel metrics specifically tailored for opinion summarization.

\section*{Limitations}
\textbf{Annotation Scale} An ideal dataset should encompass a substantial number of the following components: 1) model outputs, 2) annotations, and 3) instances. However, prior research works \cite{10.1162/tacl_a_00373, gao-wan-2022-dialsummeval} have demonstrated that an increase in model outputs and annotators typically leads to a disproportionate rise in construction time. Consider the example of annotation, where achieving the desired consensus among $n$ annotators necessitates conducting tests or re-annotations approximately $\frac{n(n-1)}{2}$ times, exhibiting a time complexity of $O(n^2)$. Consequently, to ensure high-quality annotations, we employ two annotators, meanwhile, carrying out annotations on Yelp to maximize the quantity of chosen models (14) and annotated instances (100).


\section*{Ethics Statement}
The annotators are paid 8 dollars per hour, which is above the local minimum wage, and their personal information is removed from the dataset.





\appendix
\section{A Survey of Automatic Metrics in Opinion Summarization Papers}\label{app_0}
We surveyed 21 papers from 2018 to 2023 on opinion summarization published in top NLP/AI conferences and journals: ACL \cite{isonuma-etal-2019-unsupervised, amplayo-lapata-2020-unsupervised, brazinskas-etal-2020-unsupervised, suhara-etal-2020-opiniondigest, wang-wan-2021-transsum, bhaskar2023prompted, hosking-etal-2023-attributable}, EMNLP \cite{angelidis-lapata-2018-summarizing, brazinskas-etal-2020-shot, amplayo-etal-2021-aspect, iso-etal-2021-convex-aggregation, brazinskas-etal-2021-learning}, NAACL \cite{brazinskas-etal-2022-efficient}, EACL \cite{elsahar-etal-2021-self, amplayo-lapata-2021-informative}, TACL \cite{10.1162/tacl_a_00366, isonuma-etal-2021-unsupervised}, ICML \cite{pmlr-v97-chu19b}, AAAI \cite{Li_Li_Zong_2019, Zhao_Chaturvedi_2020, Amplayo_Angelidis_Lapata_2021}. We find that the majority of papers report ROUGE-1/2/L results as the assessment of model performances, and only 4 papers \cite{elsahar-etal-2021-self, brazinskas-etal-2022-efficient, bhaskar2023prompted, hosking-etal-2023-attributable} introduce new metrics (e.g., Perplexity, BERTScore, and QA-based metrics) in addition to ROUGE as alternative evaluation methods.

\section{List of Selected Models}\label{app_list_of_selected_models}
We introduce the 14 models we selected from 4 categories: statistically-based, task-agnostic, task-specific, and zero-shot.

\textcolor{red}{\textit{\textbf{Statistically-Based Models}}}

\textbf{LexRank}$^{\textsc{Ext}}$ \cite{10.5555/1622487.1622501} is an extractive summarizer based on a PageRank-alike algorithm. By constructing a network where sentences are treated as nodes, the model selects important reviews as the output summary. We use the implementation at \url{https://github.com/crabcamp/lexrank}.

\textbf{Opinosis}$^{\textsc{Ext}}$ \cite{ganesan-etal-2010-opinosis} is a graph-based model that extracts salient reviews as the predicted summary. It connects sentences in a graph based on Part-Of-Speech (POS) tagging and selects reviews based on their redundancies\footnote{We use the \texttt{flair} toolkit \cite{akbik2019flair} for POS tagging and repeat the reviews 2-3 times to satisfy the requirement that input sentences should be $\geq 60$.}. We use the implementation at \url{https://github.com/kavgan/opinosis-summarization}.

\textbf{BertCent}$^{\textsc{Ext}}$ \cite{Amplayo_Angelidis_Lapata_2021} is a variant of the Centroid model \cite{RADEV2004919} that uses BERT embeddings to summarize. We use the resources at \url{https://github.com/rktamplayo/PlanSum}.

\textcolor{red}{\textit{\textbf{Task-Agnostic Models}}}\footnote{All the models in this category are self-implemented.}

\textbf{BART}$^{\textsc{Abs}}$ \cite{lewis-etal-2020-bart} is a PLM that uses a denoising objective to recover the original texts from random masks. We choose BART-Large as the summarizer.

\textbf{T5}$^{\textsc{Abs}}$ \cite{JMLR:v21:20-074} is trained in a unified framework where different tasks are united within a ``text-to-text'' objective. We choose T5-Base as our backbone.

\textbf{PEGASUS}$^{\textsc{Abs}}$ \cite{pmlr-v119-zhang20ae} is a PLM designed for abstractive summarization. Through sentence masking and reconstruction, it is sensitive to contexts and thus capable to generate informative summaries. We choose PEGASUS-Large as our summarization model.

\textcolor{red}{\textit{\textbf{Task-Specific Models}}}

\textbf{COOP}$^{\textsc{Abs}}$ \cite{iso-etal-2021-convex-aggregation} is an aggregation framework inspired by convex optimization which learns to summarize by maximizing word overlaps between inputs and outputs. Specifically, we choose \texttt{BiMeanVAE} with COOP as the summarizer due to its superior performance. We use the resources at \url{https://github.com/megagonlabs/coop}.

\textbf{CopyCat}$^{\textsc{Abs}}$ \cite{brazinskas-etal-2020-unsupervised} is based on multi-layer variational auto-encoders and summarizes based on the latent encodings of reviews. We use the resources at \url{https://github.com/abrazinskas/Copycat-abstractive-opinion-summarizer}.

\textbf{DenoiseSum}$^{\textsc{Abs}}$ \cite{amplayo-lapata-2020-unsupervised} disturbs the input reviews by introducing noises at the segment level and the document level, and learns to summarize from denoising. We use the resources at \url{https://github.com/rktamplayo/DenoiseSum}.

\textbf{MeanSum}$^{\textsc{Abs}}$ \cite{pmlr-v97-chu19b} is a model based on auto-encoders and learns to summarize by recovering the average encodings of reviews. We use the resources at \url{https://github.com/sosuperic/MeanSum}.

\textbf{OpinionDigest}$^{\textsc{Abs}}$ \cite{suhara-etal-2020-opiniondigest} is trained by reconstruction and can perform controllable summarization over aspects and sentiments. We use the resources at \url{https://github.com/megagonlabs/opiniondigest}.

\textbf{PlanSum}$^{\textsc{Abs}}$ \cite{Amplayo_Angelidis_Lapata_2021} tackles the unsupervised challenge via content planning, which enhances relevance in the pseudo \{reviews, summary\} pairs to construct a better training set. We use the resources at \url{https://github.com/rktamplayo/PlanSum}.

\textbf{RecurSum}$^{\textsc{Abs}}$ \cite{isonuma-etal-2021-unsupervised} is based on variational auto-encoders where summaries are generated layer-wisely. We use the resources at \url{https://github.com/misonuma/recursum}.

\textcolor{red}{\textit{\textbf{Zero-Shot Models}}}

\textbf{GPT-3.5}$^{\textsc{Abs}}$ has shown competitive abilities to perform zero-shot opinion summarization \cite{bhaskar2023prompted}. We choose \texttt{text-davinci-003} as the backbone and set the temperature to 0 while keeping the other parameters as their default.

\section{List of Evaluation Metrics}\label{app_list_of_eval_metrics}
We choose 26 metrics to evaluate their effectiveness in opinion summarization, and categorize them into non-GPT and GPT-based, depending on whether they are built upon GPTs. 

\textcolor{red}{\textit{\textbf{Non-GPT Metrics}}}\footnote{For \textbf{BLEU}, \textbf{METOR}, \textbf{EmbeddingAverage}, \textbf{VectorExtrema}, and \textbf{GreedyMatching}, we use the implementation at \url{https://github.com/Maluuba/nlg-eval}.}

\textbf{ROUGE} \cite{lin-2004-rouge} measures the n-gram overlaps between the candidate and a set of references, and is popularly used in summarization tasks. We use the implementation at \url{https://github.com/Diego999/py-rouge}.

\textbf{BLEU} \cite{10.3115/1073083.1073135} is the primary metric for machine translation. It focuses on precision and evaluates by computing n-gram overlaps between a candidate and a reference.

\textbf{METOR} \cite{banerjee-lavie-2005-meteor} measures the alignment between a candidate and a set of references by mapping unigrams.

\textbf{TER} \cite{snover-etal-2006-study} is a metric that computes the ratio between the number of edits that convert the candidate into a reference and the average number of words in references. We use the implementation at \url{https://github.com/mjpost/sacrebleu}.

\textbf{ChrF} \cite{popovic-2015-chrf} is a metric that measures the token-level n-gram overlaps between a candidate and a reference. We use the implementation at \url{https://github.com/m-popovic/chrF}.

\textbf{BERTScore} \cite{bert-score} is a metric that evaluates a candidate and a reference with their similarity based on word-level BERT embeddings. We use the implementation at \url{https://github.com/Tiiiger/bert_score}.

\textbf{BARTScore} \cite{NEURIPS2021_e4d2b6e6} measures the quality of the target text by its generation probability conditioned on the source text. We use the implementation at \url{https://github.com/neulab/BARTScore}.

\textbf{BLANC} \cite{vasilyev-etal-2020-fill} is a reference-free metric based on the assumption that summaries with quality are helpful for understanding the input documents, and evaluates by reconstructing the masked texts. We use the implementation at \url{https://github.com/PrimerAI/blanc}.

\textbf{BLEURT} \cite{sellam-etal-2020-bleurt} is based on BERT and trained with scores from different metrics as the supervision signals for evaluation. We use the implementation at \url{https://github.com/google-research/bleurt}.

\textbf{InfoLM} \cite{Colombo_Clavel_Piantanida_2022} generates distributions based on the masked word probability of texts and evaluates by calculating the similarity between the distributions of the candidate and the reference. We choose \texttt{ab-div} as the metric due to its superior performance. We use the implementation at \url{https://github.com/PierreColombo/nlg_eval_via_simi_measures}.

\textbf{BaryScore} \cite{colombo-etal-2021-automatic} is a metric that measures the similarity between a candidate and a reference based on their Wasserstein distance. We use the implementation at \url{https://github.com/PierreColombo/nlg_eval_via_simi_measures}. 

\textbf{MoverScore} \cite{zhao-etal-2019-moverscore} measures the n-gram semantic distance between a candidate and a reference based on BERT embeddings. We use the implementation at \url{https://github.com/AIPHES/emnlp19-moverscore}.

\textbf{Sentence Mover’s Similarity} \cite{clark-etal-2019-sentence} generalizes Word Mover's Distance \cite{pmlr-v37-kusnerb15} and evaluates the candidate with its distance to the reference. We consider two types of embeddings, namely, ELMo \cite{peters-etal-2018-deep} and GLoVe \cite{pennington-etal-2014-glove}. We use the implementation at \url{https://github.com/eaclark07/sms}.

\textbf{EmbeddingAverage} \cite{EmbeddingAvr} computes the cosine similarity between the embeddings of the candidate and the reference, where the average embedding of words is treated as the sentence-level embedding.

\textbf{VectorExtrema} \cite{forgues2014bootstrapping} is a metric that computes similarities based on sentence-level embeddings, which is constructed by taking the extreme value at each dimension from the embeddings of the words in a sentence.

\textbf{GreedyMatching} \cite{rus-lintean-2012-comparison} calculates the similarity by comparing words from the candidate and the reference with a greedy matching algorithm.

\textbf{Perplexity}-[\texttt{PEGASUS}] is a metric that uses a language model as the backbone to evaluate the generation likelihood of a sentence. We choose PEGASUS as our language model. We use the implementation at \url{https://huggingface.co/docs/transformers/perplexity}.

\textbf{Prism} \cite{thompson-post-2020-automatic} is a measurement that evaluates the candidate sentence by paraphrasing. We use the implementation at \url{https://github.com/thompsonb/prism}.

\textbf{$\textit{S}^3$} \cite{peyrard-etal-2017-learning} is a model-based metric trained to aggregate scores from different metrics as the evaluation result. We use the implementation at \url{https://github.com/UKPLab/emnlp-ws-2017-s3}.

\textbf{SUPERT} \cite{gao-etal-2020-supert} is a reference-free metric that measures the semantic similarity between the candidate and a pseudo reference, which is comprised of salient sentences extracted from the source documents. We use the implementation at \url{https://github.com/yg211/acl20-ref-free-eval}.

\textbf{QAFactEval} \cite{fabbri-etal-2022-qafacteval} is a QA-based metric focusing on evaluating factual consistency, which measures fine-grained answer overlap between the source
and summary. We use the implementation at \url{https://github.com/salesforce/QAFactEval}.

\textbf{QuestEval} \cite{scialom-etal-2021-questeval} is a metric that views text evaluation as a QA task and generates questions from both the source document and the candidate itself. We use the implementation at \url{https://github.com/ThomasScialom/QuestEval}.

\textbf{SummaQA} \cite{scialom-etal-2019-answers} is a QA-based metric that generates questions from source documents and treats the candidate sentence as the answer to evaluate its quality. We use the implementation at \url{https://github.com/ThomasScialom/summa-qa}.

\textbf{SummaC} \cite{10.1162/tacl_a_00453} is a lightweight metric that evaluates factual consistency using Natural Language Inference (NLI) models. We choose the SummaC$_\text{Conv}$ model as the backbone and use the implementation at \url{https://github.com/tingofurro/summac}.

\textcolor{red}{\textit{\textbf{GPT-Based Metrics}}}\footnote{All the metrics in this category are self-implemented.}

\textbf{Perplexity}-[\texttt{GPT-2}] uses GPT-2 as the backbone to evaluate the generation likelihood of a sentence.

\textbf{ChatGPT} \cite{gao2023humanlike} has shown great potential to perform human-alike evaluation. We choose \texttt{gpt-3.5-turbo} as our backbone and evaluate each summary independently. 

\textbf{G-Eval} \cite{liu2023geval} is a GPT-based metric that generates \textit{Chain of Thought} (CoT) to improve its reasoning ability when evaluating texts, and there are two variants of it. 

\textbf{G-Eval}-[\texttt{text-ada-001}] weights a set of predefined scores with their generation probability conditioned on the instructions and CoT, and we use \texttt{text-ada-001} as the backbone model. We evaluate each dimension independently.

\textbf{G-Eval}-[\texttt{gpt-3.5-turbo}] directly gives integer scores based on the instructions and CoT. We choose \texttt{gpt-3.5-turbo} as the scoring model and rate each dimension independently.

\section{Prompts for ChatGPT}\label{app_prompts_of_chatgpt}
\begin{figure}[t]
    \includegraphics[width=0.5\textwidth]{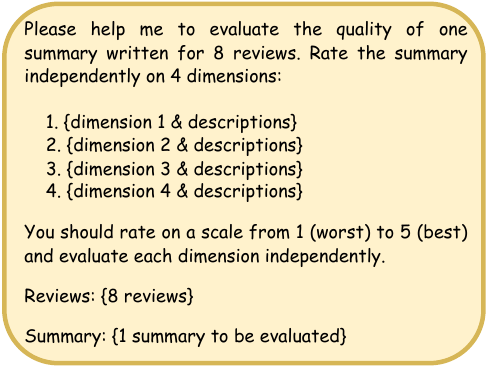}
    \caption{The prompt for \textbf{ChatGPT} \cite{gao2023humanlike}.} 
    \label{fig:chatgpt_prompt}
\end{figure}

The prompt for \textbf{ChatGPT} is shown in Figure~\ref{fig:chatgpt_prompt}. 
\section{Prompts and CoT for G-Eval}\label{app_prompts_and_CoT_of_GEval}
\begin{figure}[t]
    \includegraphics[width=0.5\textwidth]{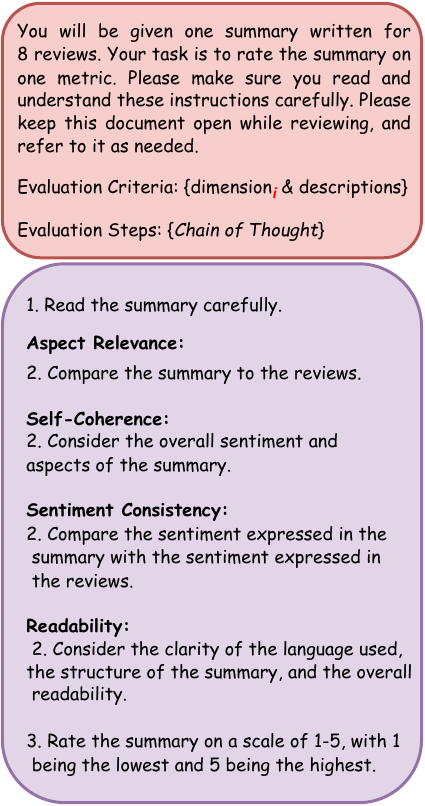}
    \caption{The prompt used in G-Eval \cite{liu2023geval} and the generated CoTs for 4 dimensions, where the differences among the CoTs for different dimensions are the descriptions for step 2.} 
    \label{fig:geval_all}
\end{figure}

The prompt for \textbf{G-Eval} and the generated CoTs conditioned on the prompt for the 4 dimensions are shown in Figure ~\ref{fig:geval_all}.
\section{A Discussion on the Choice of Dataset}\label{discussion_choice_dataset}
We show the statistics of available summaries on the two popularly used datasets in opinion summarization in Table~\ref{tab:choice_of_datasets}, where ``outputs-only'', ``checkpoint-only'', and ``both'' stand for there are only outputs publicly available, only model checkpoint publicly available, and both outputs and model checkpoint publicly available. The Amazon \cite{brazinskas-etal-2020-unsupervised} dataset is adapted from the Amazon product review dataset \cite{10.1145/2872427.2883037}, and contains 32 instances in its test set. Compared with the Amazon \cite{brazinskas-etal-2020-unsupervised} dataset, we chose Yelp \cite{pmlr-v97-chu19b} based on the following two reasons: 1. the total number of available task-specific models on Yelp (7) is larger than that of Amazon (6); 2. the total number of available instances to be annotated on Yelp (100) is larger than that of Amazon (32), which matches the annotation sizes of previous works \cite{bhandari-etal-2020-evaluating, 10.1162/tacl_a_00373, gao-wan-2022-dialsummeval}.
\begin{table*}[t]
\centering
\begin{tabular}{cc}
\hline
\multicolumn{2}{c}{Amazon \cite{brazinskas-etal-2020-unsupervised}}\\
\hline
 outputs-only & PASS \cite{oved-levy-2021-pass}, PlanSum \cite{Amplayo_Angelidis_Lapata_2021} \\
 checkpoint-only & COOP \cite{iso-etal-2021-convex-aggregation}  \\
 \multirow{2}{*}{ both} & AdaSum \cite{brazinskas-etal-2022-efficient}, CopyCat \cite{brazinskas-etal-2020-unsupervised}\\
 & RecurSum \cite{isonuma-etal-2021-unsupervised} \\
 \hline
 \multicolumn{2}{c}{Yelp \cite{pmlr-v97-chu19b}}\\
 \hline
 outputs-only & DenoiseSum \cite{amplayo-lapata-2020-unsupervised}, PlanSum \cite{Amplayo_Angelidis_Lapata_2021} \\
\multirow{2}{*}{checkpoint-only} & MeanSum \cite{pmlr-v97-chu19b}, COOP \cite{iso-etal-2021-convex-aggregation}  \\
& OpinionDigest \cite{suhara-etal-2020-opiniondigest}\\
 both & CopyCat \cite{brazinskas-etal-2020-unsupervised}, RecurSum \cite{isonuma-etal-2021-unsupervised} \\
\hline
\end{tabular}
\caption{\label{tab:choice_of_datasets} The available task-specific models of Amazon \cite{brazinskas-etal-2020-unsupervised} and Yelp \cite{pmlr-v97-chu19b}.}
\end{table*}
\section{The Detailed Annotation Process}\label{app_detailed_annotation_process}
\begin{figure*}[h]
    \includegraphics[width=\textwidth]{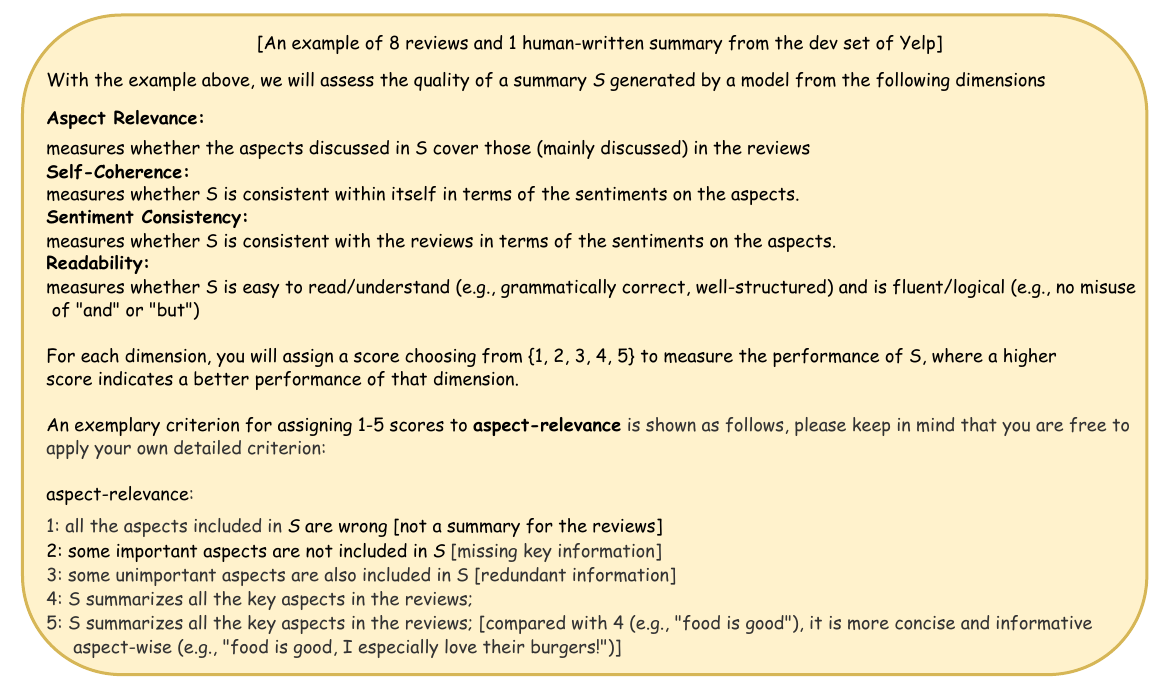}
    \caption{The guidelines for the annotation, with key information shown (we omit the example from the dev set of Yelp due to limited spaces).} 
    \label{fig:guidelines}
\end{figure*}

The annotation guideline is shown in Figure ~\ref{fig:guidelines}. After reading the guideline, the annotators are asked to conduct pilot annotations to have a better understanding of the task and are encouraged to ask questions to gain feedback. We follow \citet{alex-etal-2010-agile} to conduct agile annotation, where the annotation scheme evolves over time; thus, ensuring high annotation quality and early correction of potential mistakes. Specifically, after the $i$-th round of annotation is finished, we evaluate the annotation agreement of each batch using Cohen's $\kappa$, and batches with an agreement score less than 0.61 will later be annotated again in the $i+1$ round. After one round of annotation is finished, as the annotators become more experienced with the task, they are allowed to discuss issues related to the existing guideline and make potential refinements to it. During the entire annotation process, the annotators are promptly assisted by the authors, and are strictly forbidden to exchange ideas on giving which specific score to avoid false agreement.

\section{Interpretation of Cohen's $\kappa$ and Agreement under Other Measurements}\label{app_interpretation_of_cohens_kappa}
The interpretation of Cohen's $\kappa$ is shown in Table~\ref{tab:interpretation_of_cohens_kappa}. The annotation agreement of the final annotations calculated with Fleiss' $\kappa$ and Krippendorff’s $\alpha$ is shown in Table~\ref{tab:agreement_of_other_measurements}.

\section{A Discussion on Pearson's r}\label{app_discussion_on_pearsons_r}
Although \citet{bhandari-etal-2020-evaluating} and \citet{gao-wan-2022-dialsummeval} use Pearson's r to measure the correlations between automatic metrics and human annotations, we argue it does not apply in our case. Since Pearson's r assumes the two variables X and Y to be measured are normally distributed, we test the normality of different metrics and dimensions using \texttt{scipy.stats.shapiro}, and report the results in Table~\ref{tab:discussion_on_pearsons_r}. It is clear that only a few metrics and dimensions pass the test, which suggests that the correlations under Pearson's r only hold between certain metrics and dimensions; thus, we follow \citet{10.1162/tacl_a_00373} and adopt Kendall's $\tau$, which is a non-parametric method that does not make any assumptions on the distributions of variables.
\begin{table}[t]
\centering
\begin{tabular}{cc}
\hline
\textbf{Value of $\kappa$} & \textbf{Level of Agreement}\\
\hline
 $\leq 0$ & None \\
 $0.10\sim 0.20$ & Slight  \\
 $0.21\sim 0.40$ & Fair \\
 $0.41\sim 0.60$ & Moderate \\
 $0.61\sim 0.80$ & Substantial \\
 $0.81\sim 0.99$ & Almost Perfect \\
 $1$ & Prefect \\
\hline
\end{tabular}
\caption{\label{tab:interpretation_of_cohens_kappa} The interpretation of Cohen's $\kappa$}
\end{table}
\begin{table}[htp]
\centering
\begin{tabular}{lcc}
\hline
\textbf{} & \textbf{Fleiss' $\kappa$} & \textbf{Krippendorff’s $\alpha$}\\
\hline
\textbf{Asp.Rel.} & 0.9042 & 0.9090 \\
\textbf{Sel.Coh.} & 0.7703 & 0.7818  \\
\textbf{Sen.Con.} & 0.8250 & 0.8337 \\
\textbf{Readability} & 0.7700 & 0.7815 \\
\hline
\end{tabular}
\caption{\label{tab:agreement_of_other_measurements} The annotation agreement for each dimension.}
\end{table}
\begin{table}[t]
\centering
\resizebox{\columnwidth}{!}{
\begin{tabular}{lcc}
\hline
\textbf{metric} & \textbf{statistic} & \textbf{p-value}\\
\hline
\textbf{ROUGE-2}& 0.978 & 0.959\\
\textbf{BLEU-1}& 0.882 & 0.062\\
\textbf{BLEU-2}& 0.970 & 0.875\\
\textbf{BLEU-3}& 0.957 & 0.680\\
\textbf{BLEU-4}& 0.897 & 0.101\\
\textbf{BERTScore$_{precision}$} & 0.925 & 0.256 \\
\textbf{BERTScore$_{f1}$} & 0.911 & 0.164\\
\textbf{BARTScore$_{ref\rightarrow hyp}$} & 0.878 & 0.055 \\
\textbf{BARTScore$_{rev\rightarrow hyp}$} & 0.933 & 0.339 \\
\textbf{SMS$_{ELMo}$} & 0.937 & 0.381\\
\textbf{SMS$_{GLoVe}$} & 0.914 & 0.181\\
\textbf{GreedyMatching}& 0.943 & 0.463\\
\textbf{PPL}-[\texttt{PEGASUS}] & 0.880 & 0.058\\
\textbf{Prism}& 0.904 & 0.130\\
\textbf{$\textit{S}^3_{responsiveness}$} & 0.881 & 0.060\\
\textbf{SUPERT}& 0.959 & 0.703\\
\textbf{QuestEval}& 0.907 & 0.140\\
\hline
\textbf{G-Eval}-[\texttt{text-ada-001}]-\textbf{Sel.Coh.} & 0.935 & 0.353\\
\textbf{G-Eval}-[\texttt{text-ada-001}]-\textbf{Sen.Con.} & 0.898 & 0.105\\
\textbf{G-Eval}-[\texttt{text-ada-001}]-n-\textbf{Asp.Rel.} & 0.938 & 0.388\\
\textbf{G-Eval}-[\texttt{text-ada-001}]-n-\textbf{Sen.Con.} & 0.882 & 0.061\\
\textbf{G-Eval}-[\texttt{text-ada-001}]-n-\textbf{Read.} & 0.918 & 0.204\\
\textbf{G-Eval}-[\texttt{gpt-3.5-turbo}]-\textbf{Read.} & 0.912 & 0.168\\
\textbf{G-Eval}-[\texttt{gpt-3.5-turbo}]-\textbf{Sel.Coh.} & 0.894 & 0.092\\
\textbf{G-Eval}-[\texttt{gpt-3.5-turbo}]-\textbf{Sen.Con.} & 0.922 & 0.234\\
\textbf{ChatGPT}-[\texttt{gpt-3.5-turbo}]-\textbf{Sen.Con.} & 0.884 & 0.066\\
\textbf{ChatGPT}-[\texttt{gpt-3.5-turbo}]-\textbf{Read.} & 0.918 & 0.207\\
\hline
\textbf{Aspect Relevance} & 0.887 & 0.074\\
\textbf{Sentiment Consistency} & 0.951 & 0.578\\
\textbf{Readability} & 0.959 & 0.707\\
\hline
\end{tabular}
}
\caption{\label{tab:discussion_on_pearsons_r} The Shapiro-Wilk test \cite{eb32428d-e089-3d0c-8541-5f3e8f273532} results for different metrics and dimensions. The null hypothesis is ``\textit{the data was drawn from a normal distribution}'' and it is rejected if p-value $\leq 0.05$. We report those that passed the normality test for brevity.}
\end{table}

\section{Evaluation Results of Some Metrics}
The system-level and summary-level evaluation results for \textbf{EmbeddingAverage}, \textbf{VectorExtrema}, \textbf{GreedyMatching}, \textbf{Prism}, and \textbf{$\textit{S}^3$} are shown in Table~\ref{tab:metric_eval_result_other}.
\begin{table*}[t]
\centering
\begin{tabular}{|l|cc|cc|cc|cc|}
\hline
    \textbf{} & \multicolumn{2}{c|}{\textbf{Asp.Rel.}} & \multicolumn{2}{c|}{\textbf{Sel.Coh.}} & \multicolumn{2}{c|}{\textbf{Sen.Con.}} & \multicolumn{2}{c|}{\textbf{ Read.}}\\
    \hline
    \textbf{metric} & sys & sum & sys & sum & sys & sum & sys & sum\\
    \hline
    \textbf{EmbeddingAverage} & -0.16 & 0.09 & -0.23 & 0.05 & -0.24 & -0.02 & -0.02 & 0.05 \\
    \textbf{VectorExtrema} & 0.27 & 0.12 & 0.25 & 0.11 & 0.09 & 0.04 & 0.38 & 0.12 \\
    \textbf{GreedyMatching} & 0.05 & 0.12 & 0.12 & 0.09 & -0.04 & 0.01 & 0.27 & 0.14 \\
    \textbf{Prism} & 0.01 & 0.11 & -0.01 & 0.07 & -0.07 & -0.01 & 0.11 & 0.03 \\
    \textbf{$\textit{S}^3_{responsiveness}$} & 0.03 & 0.16 & 0.01 & 0.12 & -0.04 & 0.05 & 0.04 & 0.07 \\
    \textbf{$\textit{S}^3_{pyramid}$} & -0.12 & 0.14 & -0.14 & 0.11 & -0.20 & 0.04 & -0.11 & 0.05 \\
\hline
\end{tabular}
\caption{\label{tab:metric_eval_result_other} The Kendall's $\tau$ correlations at system-level and summary-level between automatic metrics and human annotations over 4 dimensions. For \textbf{$\textit{S}^3$}, $responsiveness$ / $pyramid$ refers to training data with a LIKERT scale / a Pyramid scheme. }
\end{table*}

\section{Case Study}\label{app_case_study}
Despite the success of \textit{task-agnostic} PLMs and \textit{task-specific} models, we observe that GPT-3.5 is consistently favored by annotators\footnote{GPT-3.5 is the best system in terms of BARTScore except for extractive models, which is because BARTScore ($rev\rightarrow hyp$) favors summaries containing sentences from the reviews.} across the 4 dimensions, which is similar to the findings of \citet{bhaskar2023prompted}. In the case study presented in Table~\ref{tab:model_output_case}, it is evident that \textbf{CopyCat} provides inaccurate recommendations, while \textbf{BART} exhibits self-contradiction. In comparison to \textbf{LexRank}, \textbf{GPT-3.5} produces well-structured, concise summaries that cover a wider range of aspects. Based on these observations, we recommend that future research in opinion summarization consider the GPT family as a baseline, as their summaries tend to closely align with human evaluation across all dimensions.
\begin{table*}[ht]
\centering \small
\resizebox{\textwidth}{!}{\begin{tabular}{p{3cm}|p{3cm}|p{3cm}|p{2.25cm}}
\hline
\textbf{LexRank} & \textbf{BART} & \textbf{CopyCat} & \textbf{GPT-3.5} \\
\hline
Don't do it. I love tacos and if you do to then go somewhere else. Horchata was warm?? Carne was bland, salsa was mild as ketchup, pastor was bland, the mf bill was sooooo high & This place is great! The food is great and the service is great. The drinks are over priced but the staff is just okay. If you want a beer, great! If you want food, try some other place. Don't do it. & If you are looking for authentic Mexican food, this is the place to go. The service was great and the food was delicious. If you haven't tried it, you will not be disappointed. & This place was terrible. The service was awful, the food was bland and overpriced, and the staff was rude. Avoid this place at all costs.\\
\hline
\multicolumn{4}{p{12.5cm}}{\textbf{Reference}\,\,\, This place was awful! how do you screw up tacos? The tacos I ordered tasted old and disgusting. The staff isn't very nice either. They always seemed rushed and are in no mood to help. Place they need to fix the air conditioner in this place. I was sweltering hot!}\\
\hline
\end{tabular}}
\caption{\label{tab:model_output_case}
Summaries of \textbf{LexRank}, \textbf{BART}, \textbf{CopyCat}, and \textbf{GPT-3.5} with the reference from a test instance of Yelp.} 
\end{table*}

\end{document}